\documentclass{ceurart}

\usepackage{todonotes,xcolor,cleveref,subcaption}
\usepackage[inline]{enumitem}

\begin{document}

\copyrightyear{2023}
\copyrightclause{by the paper’s authors. Copying permitted only for private and academic purposes.}

\conference{LWDA'23: Lernen, Wissen, Daten, Analysen.
 October 09--11, 2023, Marburg, Germany}

\title{
Higher-Order DeepTrails: Unified Approach to *Trails
}

\author[1]{Tobias Koopmann}[%
email=koopmann@informatik.uni-wuerzburg.de,
]
\author[1]{Jan Pfister}[%
email=pfister@informatik.uni-wuerzburg.de
]

\author[2]{André Markus}[%
email=andre.markus@uni-wuerzburg.de
]
\author[3]{Astrid Carolus}[%
email=astrid.carolus@uni-wuerzburg.de
]
\author[2]{Carolin Wienrich}[%
email=carolin.wienrich@uni-wuerzburg.de
]

\author[1]{Andreas Hotho}[%
email=hotho@informatik.uni-wuerzburg.de
]
\address[1]{University of Würzburg, Department of Computer Science, CAIDAS, Chair for Data Science, Germany}
\address[2]{University of Würzburg, Institute Human-Computer-Media, Psychology of Intelligent Interactive Systems, Germany}
\address[3]{University of Würzburg, Institute Human-Computer-Media, Media Psychology, Germany}

\begin{abstract}
  Analyzing, understanding, and describing human behavior is advantageous in different settings, such as web browsing or traffic navigation.
  Understanding human behavior naturally helps to improve and optimize the underlying infrastructure or user interfaces.
  Typically, human navigation is represented by sequences of transitions between states.
  Previous work suggests to use hypotheses, representing different intuitions about the navigation to analyze these transitions.
  To mathematically grasp this setting, first-order Markov chains are used to capture the behavior, consequently allowing to apply different kinds of graph comparisons, but comes with the inherent drawback of losing information about higher-order dependencies within the sequences.
  To this end, we propose to analyze entire sequences using autoregressive language models, as they are traditionally used to model higher-order dependencies in sequences.
  We show that our approach can be easily adapted to model different settings introduced in previous work, namely HypTrails, MixedTrails and even SubTrails, while at the same time bringing unique advantages:
  \begin{enumerate*}
      \item Modeling higher-order dependencies between state transitions, while
      \item being able to identify short comings in proposed hypotheses, and
      \item naturally introducing a unified approach to model all settings.
  \end{enumerate*}
  To show the expressiveness of our approach, we evaluate our approach on different synthetic datasets and conclude with an exemplary analysis of a real-world dataset, examining the behavior of users who interact with voice assistants.
\end{abstract}


\maketitle

\section{Introduction}\label{sec:intro}
Understanding and describing human behavior by analysing transitions between different actions or states has been an established field of research for several years now.
It aims to study the dynamics of human behavior by analyzing sequences of user transitions over different states and applying sequential analysis techniques.
Understanding human behavior and identifying the most common patterns of interaction can lead to improvements in many aspects, for example, web site design, traffic routing, or usability of different devices.
As an exemplary use case, we will dive into the analysis of interactions with digital voice assistants like Alexa or Google Home.
These smart devices have become increasingly popular in households over the last few years, capturing and responding to voice commands, aiming to help users with their daily tasks.
Sequences of usage behavior, if systematically analyzed, can offer valuable insight into the behavioral patterns, and therefore help improve the usability of the device.

To mathematically represent these sequences, one approach is to aggregate the sequences into graph-like structures with respective transitions between states.
Based on this, approaches have been proposed which rely on first-order Markov chain models, such as HypTrails~\cite{DBLP:conf/www/SingerHHS15}, MixedTrails~\cite{DBLP:journals/datamine/BeckerLSSH17} and SubTrails~\cite{DBLP:conf/kdd/Lemmerich0SHHS16}).
Hypotheses represent intuitions about human behavior and are constructed and ranked according to how well they fit the observed data.
We argue that this aggregation does not come without limitations: mainly the usage of first-order Markov chains is unable to capture vital information about the sequence, like higher-order dependencies.
Real-life user behavior is seldom first order; consequently, we propose to model behavior explicitly as sequences and show that allowing for higher-order dependencies by default is a natural fit for this setting~\cite{tedesco2017does}.

We propose to leverage recent advances in machine learning approaches to address this setting while being able to naturally capture higher-order dependencies in human behavior.
For this, the natural choice are autoregressive language models, commonly used in Natural Language Processing.
After fitting a model to sequences of user behavior, we propose to test the ``validity'' of a hypothesis for the training data by evaluating the model's loss.
This effectively determines whether the hypotheses exhibit expected behavior with respect to the observed user actions.
Thereby we introduce an explicitly sequence-aware variation to HypTrails, MixedTrails, and SubTrails.
The latter is a setting without available hypotheses, where we show how to incorporate transition features to analyze the sequences in a self-supervised manner.

Being able to model higher-order dependencies within user interactions provides valuable insights into user behavior patterns and decision-making processes, consequently surpassing the expressiveness of previous approaches.
The insights derived from this research have implications for improving user experience, personalizing recommendations, and designing more intuitive and adaptive systems.\footnote{Our source code is available at \url{https://github.com/LSX-UniWue/DeepTrails}.}

\section{Related Work}\label{sec:related-work}
Our work is located in the intersection of two research areas:
firstly user behavior analysis from sequences or graph-structured data and secondly sequential machine learning architectures.

\paragraph{User Behavior Analysis} describes the research domain of analyzing human behavior in any kind of sequences or graphs.
The most closely related work uses hypotheses about human behavior to evaluate to which degree a certain hypothesis fits the observed transitions.
Namely HypTrails~\cite{DBLP:conf/www/SingerHHS15} uses Bayesian inferences and sets a prior according to the believed transition probabilities from the hypothesis.
The marginal likelihood for each hypothesis with respect to the observed data is calculated, and thus, the hypotheses can be ranked according to how well they fit the observed user behavior.
MixedTrails~\cite{DBLP:journals/datamine/BeckerLSSH17} analyzed heterogeneous data, allowing researchers to study sequential data with varying behaviors.
Here, each transition is manually assigned to a group, and each group can be explained with its own hypothesis.
Furthermore, Subtrails~\cite{DBLP:conf/kdd/Lemmerich0SHHS16} proposes a method to detect interpretable subgroups with exceptional transition behavior from sequential data.
These hypothesis-driven approaches were also adapted on multigraphs~\cite{DBLP:journals/ans/NoboaLSS17} by creating a first-order Markov chain from the multigraph instead of aggregated sequences.
Finally, behavioral networks can be compared using commonly used graph metrics such as centrality, graph distance, and number of triangles~\cite{DBLP:journals/corr/abs-1904-07414}.
All of these approaches aggregate the sequences to first-order Markov chains, and hence loose information about higher-order dependencies.

\paragraph{Machine Learning for Sequential Data} has been a challenging setting, primarily due to the temporal dependencies present in the data.
In comparison, traditional machine learning models, such as Random Forest~\cite{random_forsests} or Support Vector Machines~\cite{cortes1995support}, are powerful but also limited to handling data with fixed-length feature vectors.
Nowadays, sequential data is usually processed using the transformer architecture~\cite{DBLP:conf/nips/VaswaniSPUJGKP17}.
Based on this architecture, different forms of autoregressive language models were developed~\cite{radford2019language,DBLP:conf/acl/DaiYYCLS19,DBLP:conf/nips/YangDYCSL19}, which are commonly used in Natural Language Processing, where the long-range and higher-order dependencies of words and tokens are a relevant topic.
Due to their effectiveness, sequential language models have also been adapted in other areas of research, where it might not seem intuitive at first:
e.g.\ in the research domain of recommendation~\cite{DBLP:conf/cikm/SunLWPLOJ19,DBLP:conf/icdm/KangM18}, but also graph-based machine learning approaches started by embedding nodes using sequential random walks and a form of Word2Vec~\cite{DBLP:conf/kdd/PerozziAS14,DBLP:conf/www/TangQWZYM15,DBLP:conf/kdd/OuCPZ016}.

\section{Methodology}\label{sec:methodology}
This work introduces a novel methodology to analyze and describe sequential user behavior.
For this, we follow established settings as introduced in HypTrails and its follow-up extensions.
Given a set of user observations modeled as sequences, the goal is to either find the best matching hypothesis that explains the observed user behavior (HypTrails~\cite{DBLP:conf/www/SingerHHS15} \& MixedTrails~\cite{DBLP:journals/datamine/BeckerLSSH17}) or to find ``interesting'' subgroups of users that behave differently from other groups (SubTrails~\cite{DBLP:conf/kdd/Lemmerich0SHHS16}).
These existing approaches address this topic by limiting themselves to analyzing single-step transition behavior, hence breaking the observed sequences into first-order Markov chains and analyzing these using Bayesian inference.
We argue that this inherently fails to take into account the sequential nature of the data and therefore propose using sequential machine learning models to address this problem.
Specifically, we use autoregressive language models, traditionally applied to Natural Language Modeling and sequential data, based on the intuition that the models will discover and utilize higher-order dependencies.

\begin{figure}
    \centering
    \includegraphics[width=0.9\textwidth]{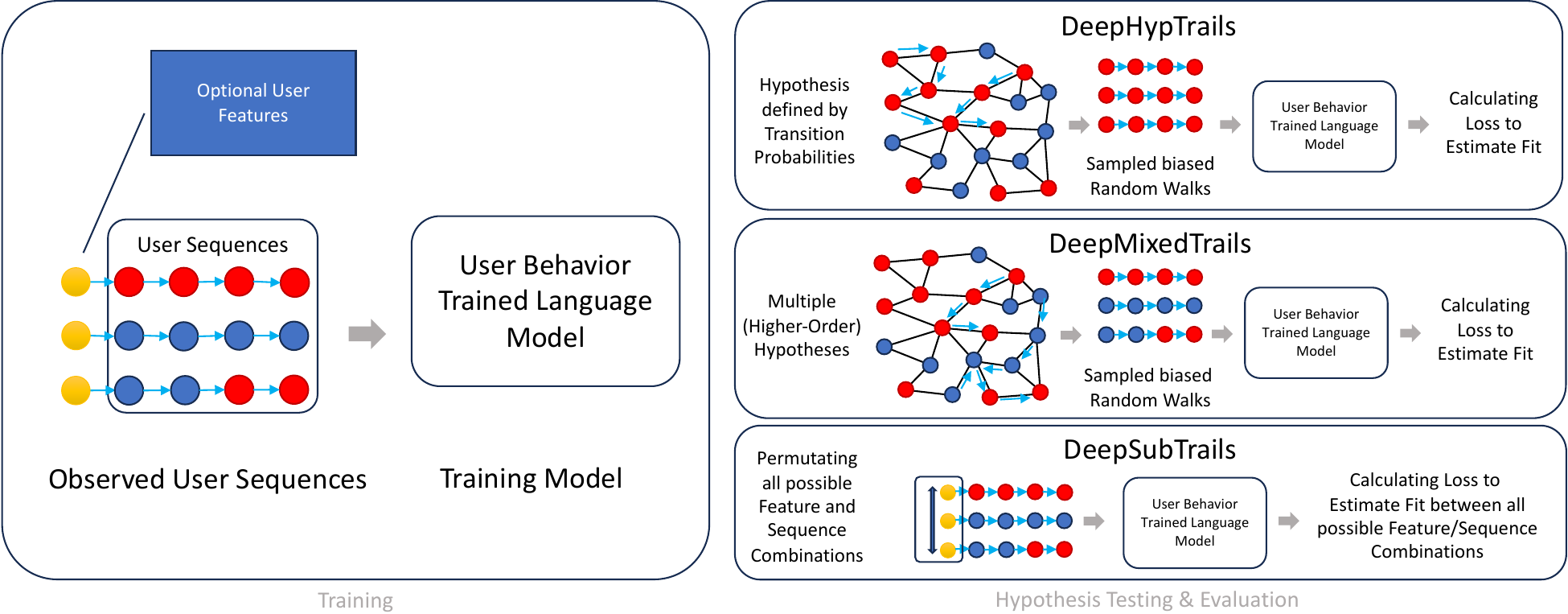}
    \caption{
    \textbf{DeepTrails}: Schematic overview of our approach. 
    We train a small language model on observed user sequences, optionally with user features. 
    Then we freeze the trained model and plug it into the respective setting:
        1. DeepHypTrails: evaluating the model on sequences generated from first-order hypothesis, or
        2. DeepMixedTrails: using sequences that contain mixed transition behavior, or 
        3. DeepSubTrails: identifying interesting subgroups of features. 
    }\label{fig:fig1}
\end{figure}

The following sections explain how we model user behavior represented as sequences with autoregressive language models, as well as measure how well a (higher-order) hypothesis matches these user sequences.
In addition to these HypTrails~\cite{DBLP:conf/www/SingerHHS15} and MixedTrails~\cite{DBLP:journals/datamine/BeckerLSSH17} settings, we also explore a setting without available hypotheses to show how to take advantage of transition features to analyze the sequences in a self-supervised manner (cf. SubTrails~\cite{DBLP:conf/kdd/Lemmerich0SHHS16}).
We can address all these settings using our language model-based approach with only minor modifications needed between the settings, as depicted in~\Cref{fig:fig1} and described in the following.
We begin by introducing our common underlying methodology in \Cref{sec:method:training,sec:method:estimating-hippos}.

\subsection{Representation of User Behavior Using Autoregressive Language Models}\label{sec:method:training}
Traditionally autoregressive language models are trained to predict the next token given the sequence of previous observations.
Mathematically, a language model can be described by the probability assigned to a sentence $x$, defined as the conditional probability over all next words $s_t$ given all previous words $s_{<t}$~\cite{DBLP:conf/nips/BengioDV00}:
$
    P(x) = \prod_{t=1}^T P(s_t \mid s_{<t}),
$
where $s_t$ is the input token at time step $t$.
We aim to exploit the fact that this is closely related to modeling sequences of user behavior by training autoregressive language models on these sequences to describe and analyze user behavior.
To this end, we model user states as input tokens by mapping every distinct state $s$ of a user's behavior to its own token, thus generating our state-vocabulary $S$.
Using teacher-forcing~\cite{DBLP:conf/nips/VaswaniSPUJGKP17} and the cross-entropy loss, we train the model to predict the next state given the sequence of all previous states (\Cref{fig:fig1}).
The model thus learns the transition probabilities for the current state $s_t$ given the entire history of previously visited states $s_{<t}$.

\subsection{Estimating How Well Hypotheses Align with the Training Data}\label{sec:method:estimating-hippos}
The core of our methodology is to measure how ``suprised'' the trained and frozen model is, when given a sequence of user behavior not seen during training.
The intuition is that a model trained on sequences of specific user behavior can adequately model sequences that were created by the same behavior, while unseen behavior is likely to be surprising to the model.
Therefore, we can identify the types of behavior the model encountered during training and those it did not.

We measure how surprised the model is when presented with a new sequence using the loss function.
For this, we freeze the model, and we calculate the positionwise cross-entropy loss of our model, between the sequence and the model's prediction, like during training, but we do not update the model.
By calculating the mean loss over the entire sequence, we obtain a measure of the model's fit for the currently evaluated sequence: a small loss indicates that the model ``expected'' this sequence, while a large loss shows that the model has not seen similar user behavior during training.
Additionally, we are able to analyze the loss per position, allowing us to provide a more in-depth analysis of the sequences, which helps us to identify where a hypothesis might be lacking - something that is not possible using the traditional HypTrails methodology.

\subsection{DeepHypTrails: Evaluating sequences on Homogeneous Observations}\label{sec:method:hyptrails}
HypTrails~\cite{DBLP:conf/www/SingerHHS15} aims to analyze sequential user behavior using hypotheses.
The setting is to identify the \emph{best matching} hypothesis for the observed data, that is, the hypothesis that best describes the observed sequences.
Since HypTrails only allows first-order dependencies, hypotheses are represented as $\left| S \right| \times \left| S \right|$ transition probability matrices, which describe the transition probabilities from one state to any other given state.

In contrast, our approach uses entire sequences, which we can generate by sampling biased random walks on the transition probability matrices (\Cref{fig:fig1}).
We then evaluate our model on these randomly generated walks as described in \Cref{sec:method:estimating-hippos}.
By averaging over all random walks - each following a specific hypothesis - we identify the hypothesis resulting in the lowest loss.

\subsection{DeepMixedTrails: Analyzing Hypotheses on Heterogeneous Observations}\label{sec:method:mixedtrails}
This extension of HypTrails allows the analysis of heterogeneous sequences.
The underlying assumption is that the sequences are not generated by a single driving force but by multiple driving forces.
Therefore, it is assumed that transition probabilities can change given the circumstances, e.g.\ after a certain amount of time, or they might even originate from entirely different groups of users.
Since the method proposed in MixedTrails~\cite{DBLP:journals/datamine/BeckerLSSH17} is strictly transition-based and relies on a first-order Markov chain, it inherently cannot distinguish if sequences
\begin{enumerate*}
    \item change behavior after a certain amount of time, i.e.\ contain higher-order dependencies, or
    \item originate from different groups, i.e.\ consist of different types of first-order behavior.    
\end{enumerate*}
Each case can be exemplary illustrated as a soccer team that plays offensive in the first half and defensive in the second or tourists and locals that have a different movement behavior in a city~\cite{DBLP:journals/datamine/BeckerLSSH17}.

Since our approach takes entire sequences into account, it is naturally able to distinguish between these two settings.
The model has no access to information on which behavior a sequence is generated by.
We rely entirely on the model to implicitly learn the different behaviors present on its own, such that the loss is low when testing hypotheses that match the training sequences, and high if the behavior was not present in training (\cref{sec:method:estimating-hippos}).
Additionally, this even allows us to combine both scenarios of MixedTrails, by training the model on sequences from different groups, where each might contain higher-order dependencies.

\subsection{DeepSubTrails: Identifying Subgroups with Interesting Behavior}\label{sec:method:subtrails}
Lastly, we use our approach in the SubTrails setting~\cite{DBLP:conf/kdd/Lemmerich0SHHS16}.
For this, instead of using predefined hypotheses to describe human behavior, we analyze observed user behavior to find unique subgroups that show ``exceptional'' transition behavior.
These subgroups exhibiting interesting transition behaviors - when compared to all transitions - are identified by attributes or features assigned to each transition.
We have to slightly adapt our approach to accommodate for this scenario, by conditioning our autoregressive model on the features that are associated with the sequence (details in \Cref{method:models}).
To this end, we feed all features to the model, and thus enable it to learn a different conditional behavior based on the given feature expressions.
For evaluation, we assess the loss across all possible combinations between features and observed sequences, thereby identifying interesting user behavior through the corresponding loss values.
A large loss indicates that the given combination of feature expression and sequence is unexpected to the model and, respectively, shows atypical sequential behavior for this feature expression.
Furthermore, we introduce a similarity measure for features: features are similar if the model assigns a similar loss for the same sequences, when conditioned with the respective feature.
This measure allows us to cluster similar feature expressions together, and thus identify similarly behaving users.
As before, our approach has the advantage of naturally capturing higher-order dependencies.

\section{Experimental Setup}\label{sec:experiments}
We create several different artificial datasets with known synthetic user behavior.
Additionally, we demonstrate the expressiveness of our approach on a real-world dataset, in which we analyze the behavior of users interacting with voice assistants.

\subsection{Generation of Synthetic Datasets}\label{sec:experiments:dataset}
We generate several synthetic datasets containing different types of user behavior that can be described or uncovered using our methodology introduced in \Cref{sec:methodology}.
To artificially generate user transitions, we define user behavior over an underlying Barabasi-Albert graph~\cite{Barabasi99emergenceScaling}.
This graph structure is scale-free, which means that the graph is densely connected in the ``center'' and less connected in the periphery.
For all synthetic datasets, we generate a graph with $n=100$ nodes and new nodes connected to $m=10$ existing nodes using preferential attachment~\cite{Barabasi99emergenceScaling}.
We divide the graph nodes equally into \emph{even} and \emph{odd} nodes, allowing us to define synthetic user behavior based on these classifications.
To model different types of behavior, we introduce differently biased random walkers that follow a predefined transition behavior based on the node categories.
For each synthetic behavior, we start a new biased random walker from each node 1000 times and generate sequences of length 20.
Notably, the model has no explicit information about the underlying graph structure or node categorization and only has access to the exhibited user observations.

\paragraph{Synthetic Data for HypTrails}
To evaluate our approach, we define different synthetic behaviors as follows: 
\begin{enumerate*}[label=(\roman*)]
    \item \textbf{even:} The walker only transitions towards even nodes
    \item \textbf{odd:} The walker only transitions towards odd nodes
    \item \textbf{random:} The walker randomly transitions towards any adjacent node
    \item \textbf{teleport:} In contrast the walker randomly teleports to any node on the graph.
\end{enumerate*}
Additionally, to show the applicability of our model to noisy data, we create matching biased probabilistic walkers for each behavior, where the walker follows the hypothesis only 90\% of the time, otherwise following the opposite behavior.
Following each of the above behaviors, we generate separate sets of sequences for observed user behavior as well as for hypotheses, which will be used for evaluation.

\paragraph{Synthetic Data for MixedTrails}\label{sec:exp:mixedtrails}
As explained in \Cref{sec:method:mixedtrails} we distinguish between two scenarios
\begin{enumerate*}[label=(\roman*)]
    \item The sequences are generated by different driving forces or
    \item transitions within a sequences are created by a changing behavior.
\end{enumerate*}
The latter scenario can be modeled by using higher-order dependencies in the creation process.
We create a single dataset containing both scenarios at the same time: a mixture of sequences originating from several differently biased random walkers, and some of them even showing varying behavior over time.
To this end, we add the following behaviors:
\begin{enumerate*}[label=(\roman*)]
    \item \textbf{first-even:} The walker only transitions to even nodes for the first half of the walk, and only odd nodes for the second half.
    \item \textbf{first-odd:} The walker only transitions as above, but in reverse.
    \item \textbf{two-odd-two-even:} The walker transitions twice to odd nodes, then twice to even nodes, and so forth. 
\end{enumerate*}
The dataset consists of 330 walks per node for each of the following behaviors: \texttt{even}, \texttt{odd} and \texttt{first even}.

\paragraph{Synthetic Data for SubTrails}
Following the SubTrails approach, we only generate observed sequences and no hypotheses, but introduce additional matching feature vectors for each sequence.
The model has access to this feature vector consisting of attributes that \emph{may} influence user behavior.
Some features explicitly correlate with the observed behavior, while some do not.

Here, our dataset consists of 250 biased random walks per node each: \texttt{even}, \texttt{odd}, \texttt{first even}, and \texttt{first odd} bias.
We add a feature vector with six binary features, where the first feature activates only if the random walker follows the \texttt{even} bias, the second feature only activates if the random walker follows an \texttt{odd} bias, etc.
The last two features are activated at random and serve as noise that the model has to learn to ignore. 
This results in a total of 16 possible feature combinations, where the last two features exhibit four potential permutations. 
The first 4 features are one-hot encoded, thereby combining the 4 permutations to each one-hot encoding, we can create 16 possible feature combinations.

\subsection{Case Study on Long-Term Study on Voice Assistant Usage}\label{sec:experiments:dataset_ass}
As a real-world case study of our methodology, we analyze a dataset containing various types of user behavior.
It was created by a long-term study in which 39 students received an Amazon Alexa or Google Home device and their usage was tracked and analyzed.
For our analysis, we represent user interactions with these devices as sequences.
To this end, we introduce 33 distinct states that are shared between all sequences.
Every state represents one type of command, e.g.\ ``playing music'' or ``asking about the news''.
Sequences are constructed from consecutive voice commands:
Every time a user interacts with the device twice within a 15-minute time window, a transition between the two types of voice commands is added for the current sequence.
Thus, we construct 217 sequences with a length between 2 and 94 and an average of 6.43.
Information about the users' perception of the voice assistants as well as other psychological features have been systematically collected using questionnaires.
We use these features, e.g.\ how lonely a user is or to which extent the user describes the voice assistant as a friend as additional input for the model.
We also use time of day and day of the week as additional metadata, leading to 150 distinct feature sets, each used by at least one user.

\subsection{Autoregressive Language Models}\label{method:models}
Our methodology requires a model that can be used to autoregressively model sequences.
Traditional language models are an obvious fit, but we also explore a significantly smaller and deterministic language model based on a Random Forest.

\paragraph{Transformer Decoder}
The first model uses a transformer decoder following the GPT architecture.
Every state in the sequence is modeled as a token.
The model is trained to predict the next state given the history of current states autoregressively using teacher forcing~\cite{DBLP:conf/nips/VaswaniSPUJGKP17}.
We use the nanoGPT implementation\footnote{\url{https://github.com/karpathy/nanoGPT}}  with a vocabulary size of ``number of states'' plus 2 (e.g.\ 100 nodes + EOS and BOS tokens for the synthetic datasets), four layers with four heads, and an embedding dimension of 16.
For our DeepSubTrails setting, we additionally adapt the model to encode a feature vector.
Categorical features are encoded as a one-hot vector, whereas numerical features are kept as is.
All features are concatenated into a vector and then consequently embedded to match the dimensionality of the token embedding.
This new token embedding replaces the BOS token, thus conditioning the model to predict a sequence with respect to the given user features.
This is comparable to user embeddings in sequential recommendation~\cite{fischer2022personalization}.

\paragraph{Random Forest-based Language Model}
As a second model, we will use a language model based on a random forest classifier and train it on our sequences.
For this, we encode each state as a one-hot vector and multiply it by an exponentially decaying weight, depending on the positional distance of the embedded token to the current token. 
This can be compared to a positional embedding.
All these weighted one-hot encoded vectors are summed up into a single vector, and thus the shape becomes independent of the sequence length.
The positional information for each state is encoded in the magnitude of this multi-hot vector.
For modeling user features, we create a vector which contains an integer per categorical feature and a float per numerical value, and concatenate it to the sequence information described above.
Thus, the model has access to the sequence and user information at the same time.

\section{Results}\label{sec:results}
In the following section, we will show the different experiments conducted, interpret the results, and show further analysis with respect to the learned embeddings.
We conduct experiments for all of the previously mentioned scenarios, namely DeepHypTrails, DeepMixedTrails, and DeepSubTrails, and additionally show the applicability of our approach in a real-world setting, where users interact with a voice assistant.

\subsection{DeepHypTrails}

\begin{figure}
    \hspace*{\fill}%
    \begin{subfigure}{0.4\textwidth}
        \includegraphics[width=\textwidth]{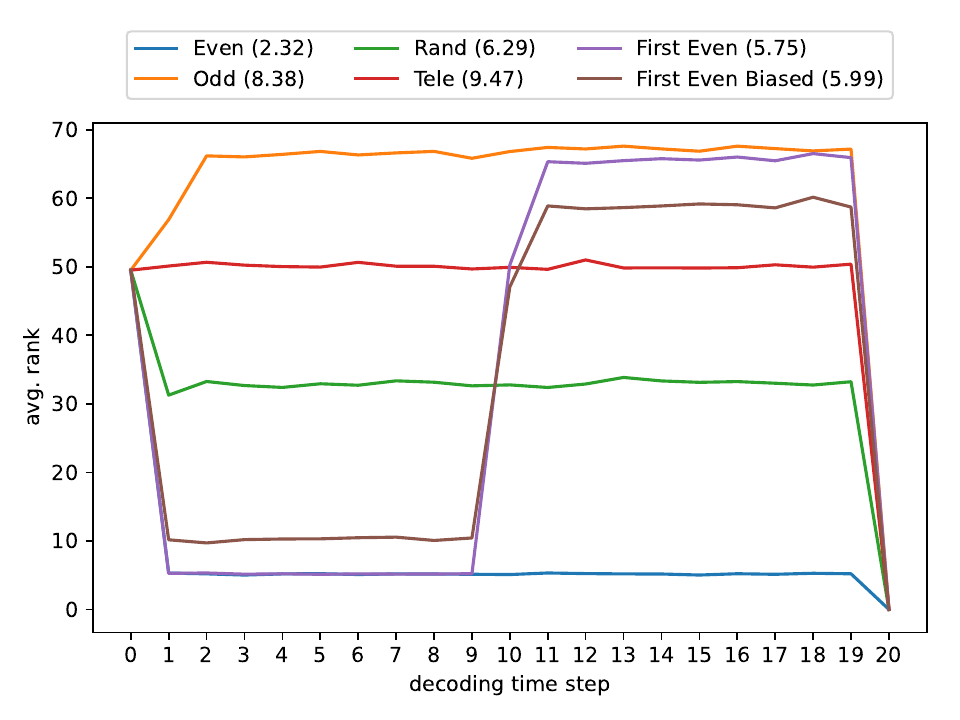}
        \caption{HypTrails setting}\label{fig:even-biased-decoding}
    \end{subfigure} \hfill%
    \begin{subfigure}{0.4\textwidth}
        \includegraphics[width=\textwidth]{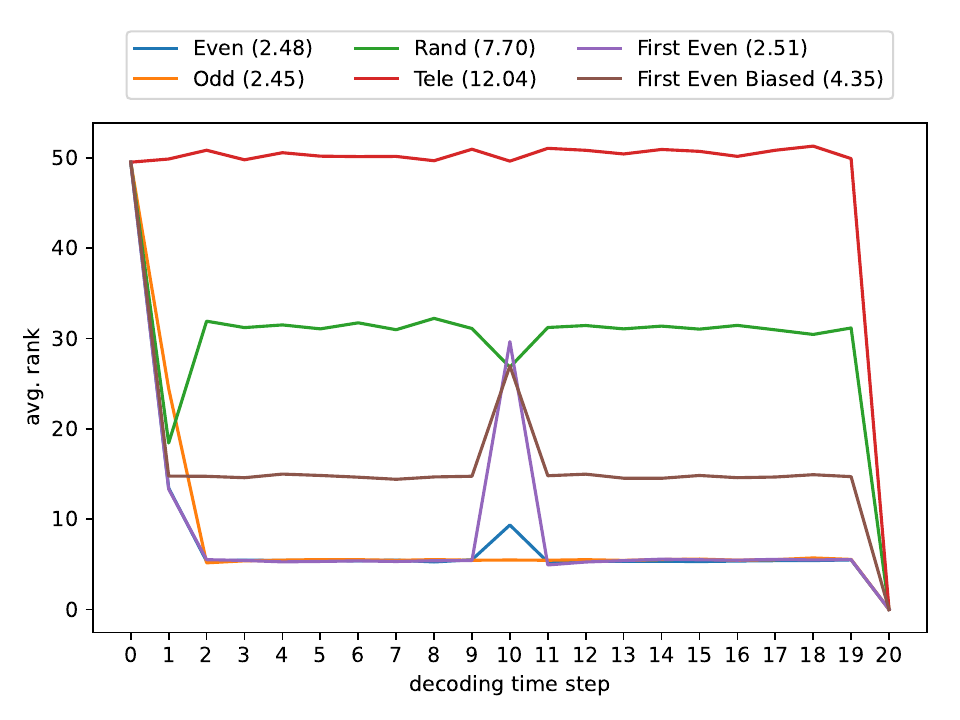}
        \caption{MixedTrails setting}\label{fig:mixed}
    \end{subfigure}%
    \caption{Applying autoregressive language models, here GPT, to sequential behavior analysis. 
    \Cref{fig:even-biased-decoding} shows the results trained on sequences with only \texttt{even} behavior and evaluates it on different hypotheses. 
    For each time step, we rank the predicted tokens by logits in descending order. 
    We plot the average rank of the target token per decoding step for each hypothesis. 
    A low rank indicates a high likelihood for the next transition according to the model.
    \Cref{fig:mixed} shows the result when training on sequences of heterogeneous behaviors, as introduced in \Cref{sec:exp:mixedtrails}.
    The number in the legend indicates the average loss over all sequences and positions for the given hypothesis.
    }
    \label{fig:hyptrails}
\end{figure}

\begin{figure}
    \hspace*{\fill}%
    \begin{subfigure}{0.4\textwidth}
        \includegraphics[width=\textwidth]{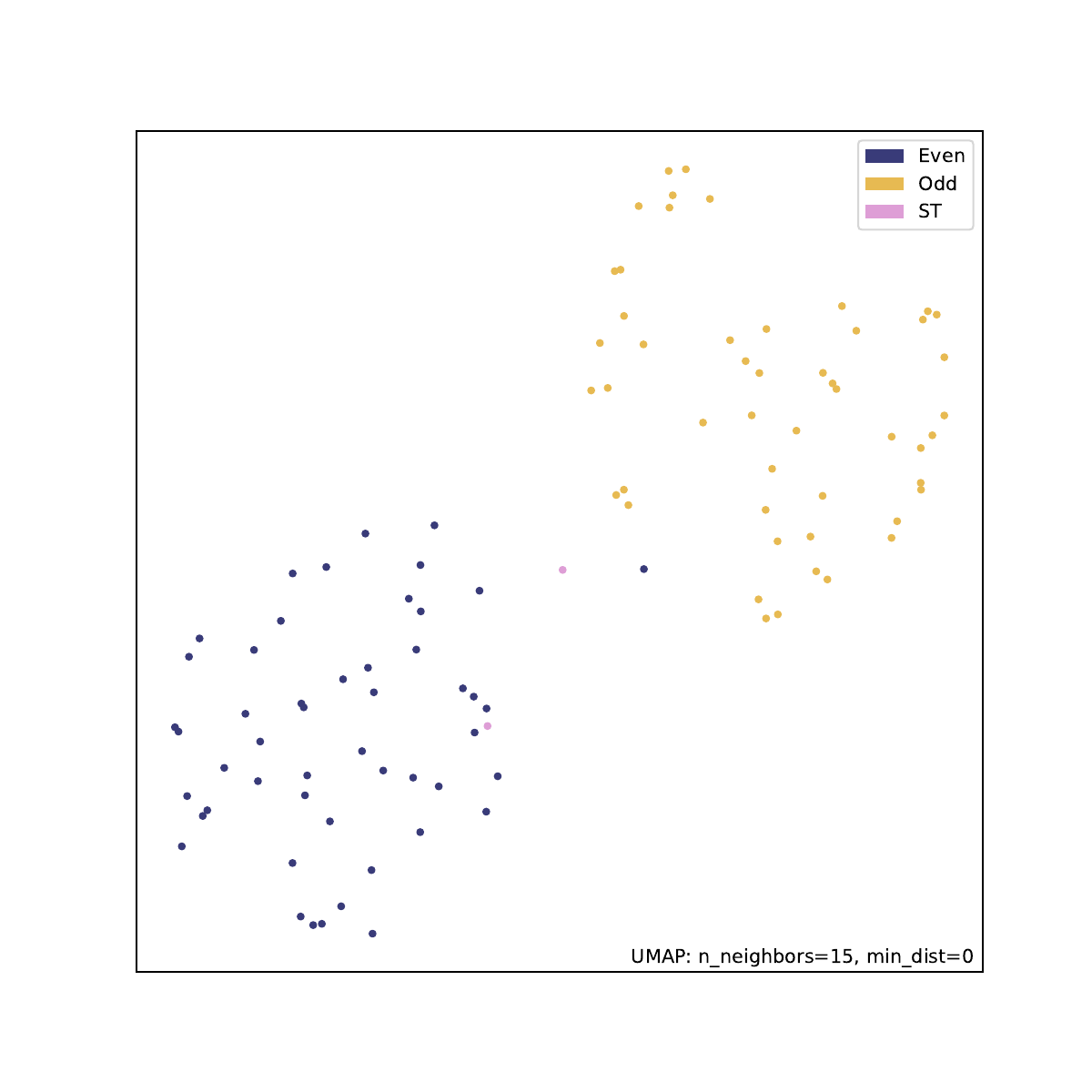}
        \caption{Analysis of Embedding space}\label{fig:first-even-biased-umap}
    \end{subfigure}\hfill%
    \begin{subfigure}{0.4\textwidth}
        \includegraphics[width=\textwidth]{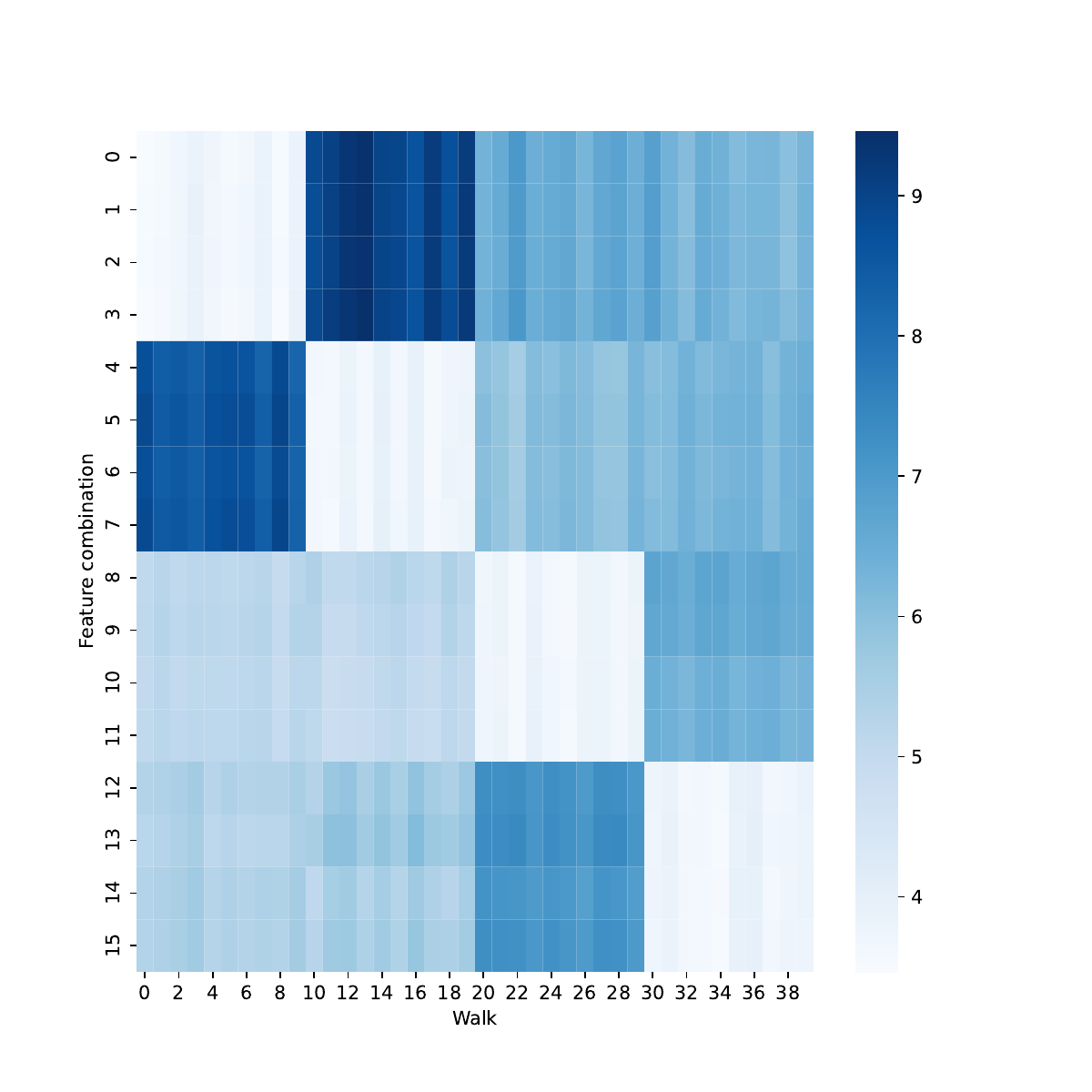}
        \caption{Heat map of loss on synthetic data}\label{fig:synheatmap}
    \end{subfigure}
    \hspace*{\fill}%
    \vskip\baselineskip
    \hspace*{\fill}%
    \begin{subfigure}{0.4\textwidth}
        \includegraphics[width=\textwidth]{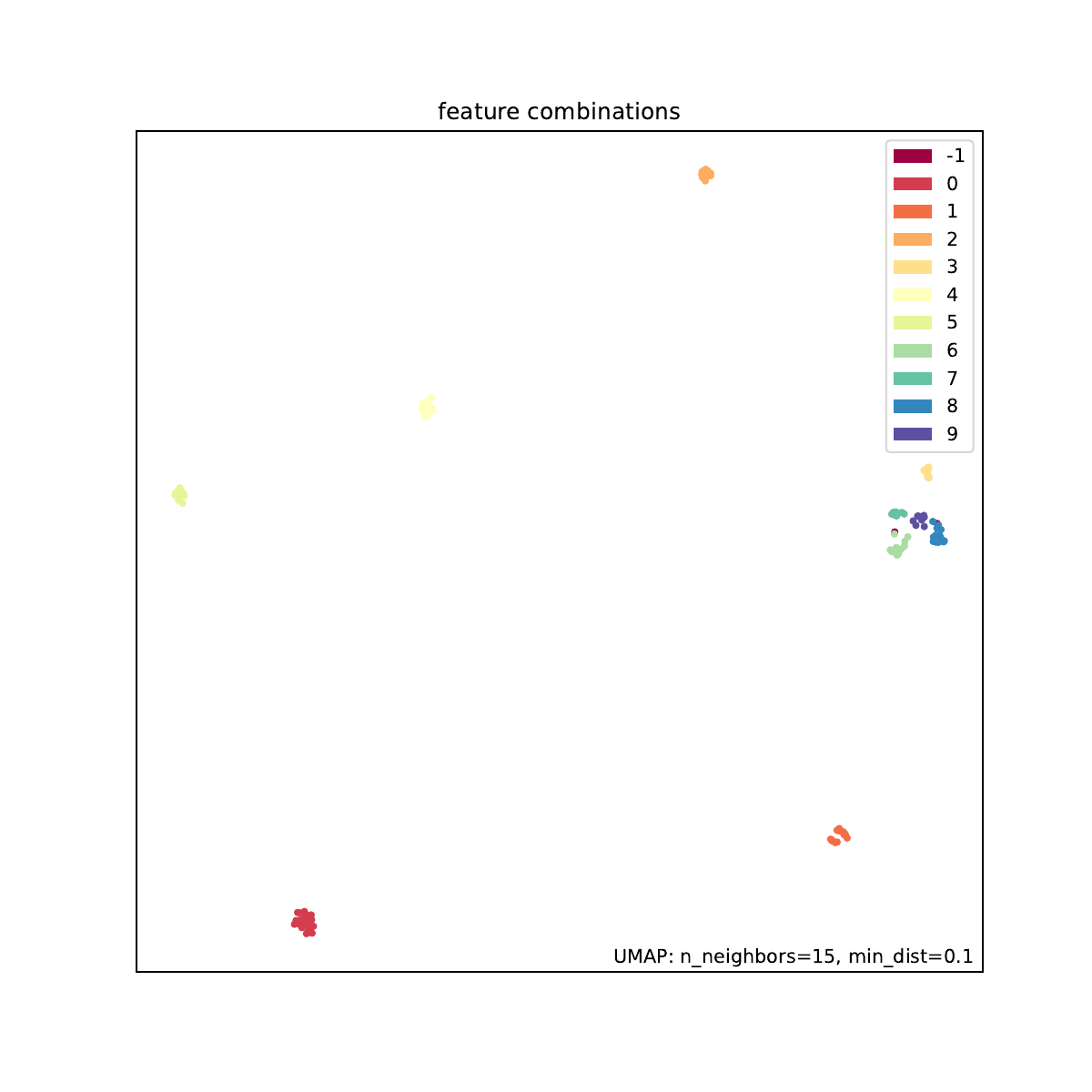}
        \caption{Feature clustering w.r.t. the probability}\label{fig:amzfeatclusters}
    \end{subfigure}\hfill%
    \begin{subfigure}{0.4\textwidth}
        \includegraphics[width=\textwidth]{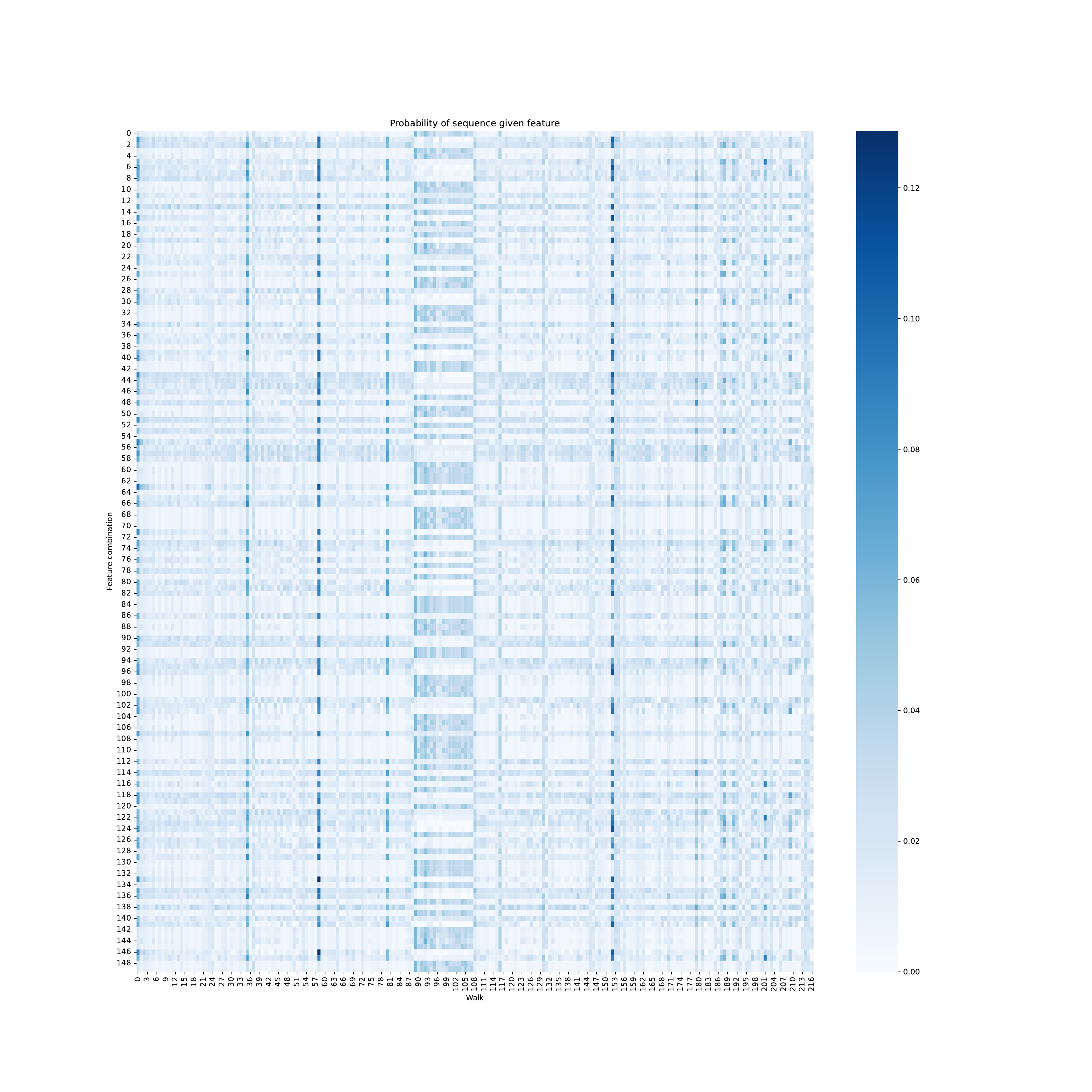}
        \caption{Heat map for Voice Assistant dataset}\label{fig:amzheatmap}
    \end{subfigure} 
    \hspace*{\fill}%
    \caption{ 
    \Cref{fig:first-even-biased-umap} shows the ability of our model to learn the graph structure and the different node types. 
    We use a UMAP visualization of node embeddings, which shows a separation between the different node types.
    \Cref{fig:synheatmap} shows a visualization of the DeepSubTrails loss scores in synthetic data. 
    The model is trained on sequences prepended with features. 
    For evaluation, we create different combinations of features and walk behavior and visualize the loss.
    \Cref{fig:amzheatmap} shows the same visualization in our real-world dataset.
    Here the x-axis lists all command sequences and the y-axis all distinct feature combinations.  
    \Cref{fig:amzfeatclusters} clusters the possible feature sets with respect to the column-wise probability given by our model from \Cref{fig:amzheatmap}. 
    }
    \label{fig:mixtrails}
\end{figure}

In the following, we evaluate the GPT model in the DeepHypTrails setting on synthetic data as previously introduced. 
The Random Forest language model leads to similar results, but we limit our analysis here to GPT due to space constraints; additional plots for the random forest model can be found in the appendix\footnote{Appendix can be found on \url{https://professor-x.de/deeptrails-arxiv}. }.
\Cref{fig:even-biased-decoding} shows the rank-wise analysis for different hypotheses of the output of our model, when trained on \texttt{even} behavior.
The x-axis shows the decoding time step, i.e.\ the position in the walk.
At each time step, we rank the tokens according to their logits in descending order.
The y-axis shows the average rank of the target token per position, where a low rank indicates a high likelihood for the next token according to the model.
All hypotheses begin with an average rank of \textasciitilde50, since we start from each of the 100 nodes in the graph equally often.
As the model has to randomly guess the first node from which the walk starts, the expected average rank for this step is 50.
Furthermore, the last node at decoding step 20 has an average rank of 1, since all synthetic sequences have a length of 20 and the model is able to predict the position of the EOS token perfectly.
Sequences created from hypotheses showing the same behavior as the observed sequences have the lowest average ranks, as expected and shown by the blue line (\texttt{even}).
The opposing behavior (orange line, \texttt{odd}) leads to average ranks above 60, which shows that the model expects these nodes to be very unlikely.
Using a random walker (green line, \texttt{rand}) as hypothesis leads to an average rank of \textasciitilde30, which is higher than a hypothesis with \texttt{even} bias, but also lower than the \texttt{odd} walker.
Naturally, transitioning at random is more similar to the observed user behavior than actively following an opposite behavior than during training.
Furthermore, a teleporting walker (red line, \texttt{tele}) stays at an average rank of \textasciitilde50 after the first step, which is expected, since this again means that the model cannot predict the next transition.
Furthermore, this shows the ability of the model to learn the graph structure, since the average rank of the \texttt{rand} hypothesis is lower, showing that the model is less surprised by a randomly sampled adjacent node than any randomly sampled node.
Finally, using sequences with higher-order dependencies in \Cref{fig:mixed}, namely \texttt{first even}, we can observe a similar rank for the first half of the sequences.
Subsequently, the average rank increases to the same level as the \texttt{odd} hypothesis after the behavior changes.
We find that \texttt{first even biased} results in comparable ranks, but with a smaller amplitude, as it effectively is a mixture of \texttt{first even} and \texttt{rand}.
This shows that our approach is capable of ranking the sequences generated with the respective behavior accordingly and is even possible to analyze sequences by position, thus indicating at which transitions a hypothesis might not be adequately explaining the data.
Furthermore, in \Cref{fig:first-even-biased-umap}, we analyze the extracted token embeddings for each state of a model trained on \texttt{first even} observations.
The UMAP plot with annotated classes shows that the model is able to neatly separate both state types in the token embedding space, indicating that the model is able to understand the different types of nodes. 
The label ``ST'' denotes the embeddings of the special tokens.

Finally, we show an ablation study in the appendix, where we applied the original HypTrails approach on data created by higher-order sequences. 
Our results show that HypTrails, due to the use of a first-order Markov chain, is not able to distinguish between the higher-order sequences and random navigation.  

\subsection{DeepMixedTrails}
In the next experiment, we use our approach in both MixedTrails settings as introduced in \Cref{sec:method:mixedtrails}.
In \Cref{fig:mixed}, we show the results for a model trained on sequences from different groups of users who also show a change in behavior over time, as explained in \Cref{sec:experiments:dataset}.
We expect hypotheses that exist in the training observations to have on average a lower rank than hypotheses not occurring in the data.
Since the model is trained on equal amounts of \texttt{even}, \texttt{odd} and \texttt{first even} behavior, all of these hypotheses lead to similarly low ranks (see \Cref{fig:mixed}).
Notice that the rank at step 1 is about 15, since the model does not know yet to which behavior the current sequence belongs to - either one of the even behaviors follows if it is an odd walk.
After time step one this is unambiguous, thus the rank decreases further.
At decoding step ten, we can observe a peak which is explainable by higher-order hypotheses \texttt{first even}.
The model cannot know whether the current hypothesis follows \texttt{even} or \texttt{first even}, therefore this spike appears for both hypotheses.
In particular, this spike is absent for the \texttt{odd} hypothesis as the model can be certain to follow this behavior known from the training data.
The average rank for the \texttt{rand} hypothesis increases after the first decoding step, since the model assumes that the subsequent steps follow the first observed time step and not random nodes.
As before, \texttt{tele} remains at a rank of \textasciitilde50.
We show additional experiments for DeepMixedTrails in the appendix.

\subsection{DeepSubTrails}
As final synthetic experiment, we analyze transitions with respect to features, instead of using hypotheses.
For this, we train our model on the synthetic dataset described in \Cref{sec:experiments:dataset}, which contains a feature vector with six categories for each walk.
For evaluation, we use 10 exemplary walks per behavior (40 total).
Walks 0 to 10 are walks containing \texttt{even} behavior, 11 to 20 contain \texttt{odd} behavior, 21 to 30 contain \texttt{first even} behavior and 31 to 40 contain \texttt{first odd} behavior. 
We sort the matching feature sets in the same ordering, where feature sets 0 to 4 have category 1 activated, all possible permutations for category 5 and 6 (which are walks containing \texttt{even} behavior), and so forth.
We combine the 16 unique feature sets with each of the 40 walks and visualize the resulting loss as a heat map in \Cref{fig:synheatmap}.

We observe the lowest loss across the diagonal, where the model correctly finds that the behavior in the walks matches the features.
The highest loss can be found in the dark blue rectangles in the upper left corners, where the feature suggests \texttt{even} walks, but the walks contain only \texttt{odd} behavior and vice versa.
Next, we can observe medium loss scores for the top right and bottom left corner, where the feature suggests changes in behavior walks, but the walks match only partially by containing only \texttt{odd} or \texttt{even} behavior.
Finally, we can observe that noise categories 5 and 6 do not have any impact:
the loss is stable across all feature combinations where only these noncorrelating categories change (feature combination 0 to 3, 4 to 7, etc.).
Feature combinations 0 through 3 iterate over all possible combinations only of categories 5 \& 6, but the loss and the other categories remain stable.
By recombining features and sequences, this allows us to estimate how well a feature set matches to each walk and therefore hypothesis and behavior. 
We add further analysis and visualizations by clustering the features and sequences in the appendix. 

\subsection{Case Study: DeepSubTrails on Voice Assistant Data}\label{sec:case_study}
The last experiment uses the DeepSubTrails methodology on our voice assistant dataset, as previously introduced (\Cref{sec:experiments:dataset_ass}) using the random-forest based language model.
\Cref{fig:amzheatmap} exhibits clear patterns in the data by showcasing which feature set best matches which sequence.
We find that several rows possess the same probability distribution across all sequences (columns), showing that these feature sets indicate similar transition behavior.
The same phenomenon can be observed when analyzing the heat map column-wise, w.r.t.\ the sequences: several sequences are similar to each other based on which sets of feature expressions they match best.
To analyse this more intuitively, in \Cref{fig:amzfeatclusters}, we use UMAP and HDBSCAN to cluster similar feature sets by interpreting the scores in each row from the heatmap as its vector representation.
Thus, we can identify ten different clusters of feature sets that behave similarly and, therefore, yield similar probabilities for the same sequences.
When analyzing these clustered feature sets, we find that commonly different feature sets are clustered together, which differ only by the ``day of the week'' feature, indicating consistent user behavior across different week days.

Overall we find that there are, in general, two types of clusters: mono-user clusters and multi-user clusters.
Six clusters consist of only a single user each, which exhibits the same behavior across different days of the week, but at similar times of the day.
For example in cluster ten, we find a cluster of feature sets all stemming from the same user, and only differing by the ``day of the week''.
The feature sets in the cluster all stem from interactions in the morning and in the associated sequences the user asks the assistant to play music, sometimes in combinations with setting a timer.
In the same spirit cluster six consists of a single user who sets \emph{multiple} alarms in the evening, often times in combination with controlling their smart-home and media-playback.
This hints at a fixed incorporation of the voice assistant in their bed-time routine.
Furthermore cluster two and three consist of different users oftentimes repeatedly asking to play or control their media in the afternoon.

Next, we exemplary analyse the clusters consisting of feature sets stemming from different users.
Naturally they are more heterogeneous in their behavior, making interpretation harder.
Nevertheless, we find clusters eight and nine, consisting of two users and four users respectively whose voice commands get misunderstood - oftentimes multiple times in a row.
This either stems from the device misunderstanding certain song titles and bands, or more commonly: trying to control the device with media controls like ``Alexa, next!'', when the current menu (e.g. within a skill) is not controllable with these commands.
At the same time, within cluster nine something similar happens but for e.g.\ Bluetooth control or the calculator.
Here it is also common that no valid or understood command is executed within a sequence, indicating that the user has given up altogether in trying to make the device perform the requested action.
Overall this shows the expressiveness of our methodology, as well that it is possible to meaningfully analyse real world datasets with it.

\section{Conclusion}
This work explores the use of autoregressive language models to analyze and describe sequential user behavior.
We train a model on observed sequences, therefore, predict user behavior present in these sequences and test hypothesis on user behavior. 
We evaluate our approach in three different settings, which are adopted from previous work.
For the first two settings, we construct (higher-order) sequences, sourced from hypotheses, each embodying a possible explanation for the observed user behavior.
Consequently, we calculate the loss of the frozen model on these generated sequences to estimate the fit of the hypothesis for the observed behavior.
In a third setting, where user features but no hypotheses are available, we can find exceptional feature sets that indicate unique user behavior.  
We show the applicability and advantages of our approach on several synthetic datasets and conclude by showing one setting on a real-world dataset, namely the usage behavior of users interacting with their voice assistants.

\section*{Acknowledgements}
This work is partially supported by the German Research Foundation (DFG) under grant number 438232455 (HydrAS) and the MOTIV research project funded by the Bavarian Research Institute for Digital Transformation~(bidt), an institute of the Bavarian Academy of Sciences and Humanities.
The authors are responsible for the content of this publication.

\bibliography{sample-ceur}
\newpage
\section{Appendix}

The appendix contains some additional analysis, which is helpful for a deeper understanding of our work and did not fit in the main manuscript.
First, we show some additional experiments for our DeepHypTrails, where we add a model trained on random transitions and one trained on biased observations. 
Then, we show the second scenario for the DeepMixedTrails approach, where the observed data contain sequences with changing behavior.
Furthermore, we will show further analysis for the DeepSubTrails approach by clustering the sequences and features. 
Finally, we show the same visualizations for our real-world setting on voice assistant commands.

\subsection{DeepHypTrails: Using biased sequences as observations}

\begin{figure}[h]
    \centering
    \hspace*{\fill}%
    \begin{subfigure}{0.4\textwidth}
        \includegraphics[width=\textwidth]{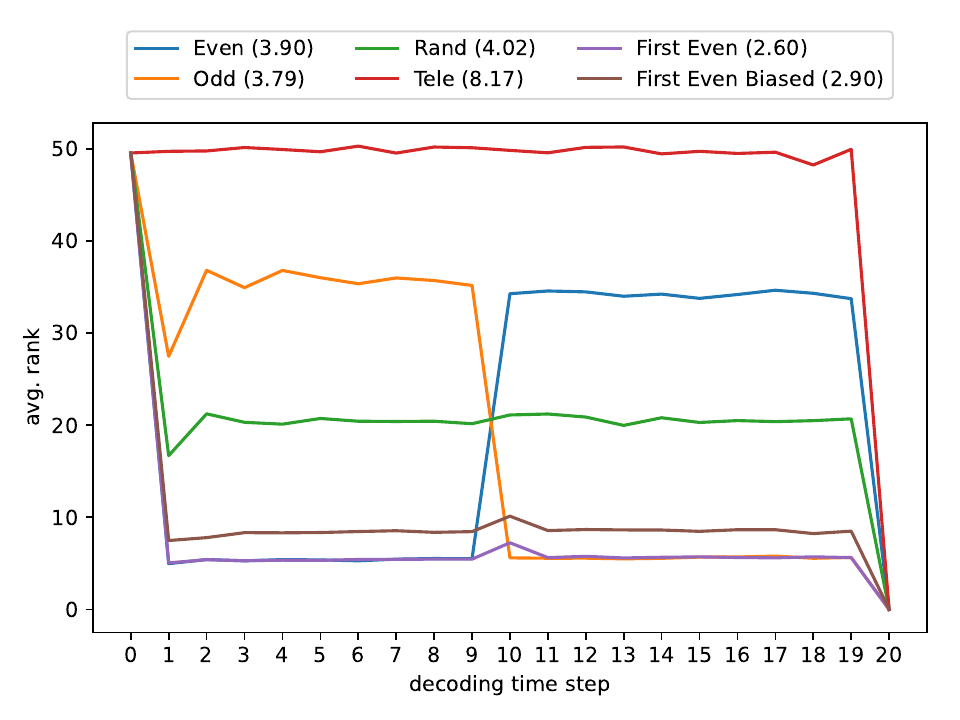}
        \caption{Model trained on biased data, namely \texttt{first even biased}}
        \label{fig:first-even-biased}
    \end{subfigure}\hfill%
    \begin{subfigure}{0.4\textwidth}
        \includegraphics[width=\textwidth]{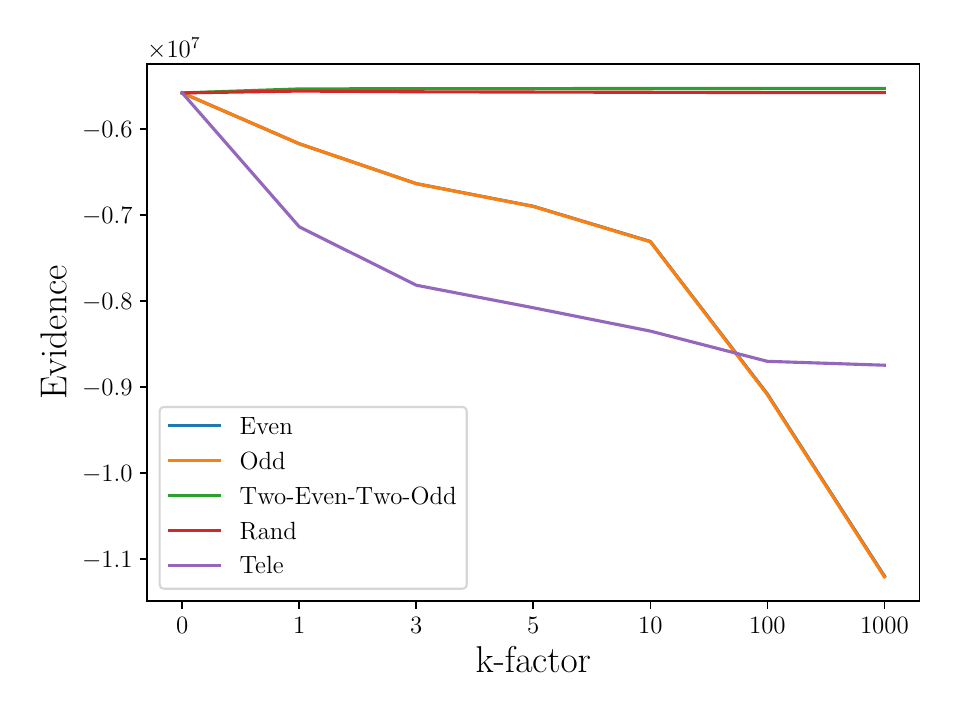}
        \caption{Applying HypTrails on higher-order sequences. }\label{fig:hyptrails-ablation}
    \end{subfigure}%
    \hspace*{\fill}%
    \caption{\Cref{fig:first-even-biased} is trained on \texttt{first-even-biased}, which means that the behavior is chosen using 90\% and 10\% the opposite behavior. 
    \Cref{fig:hyptrails} shows a HypTrails analysis on higher order sequences. 
     The y-axis shows the evidence (higher is better), and the x-axis the concentration factor. 
    The underlying data were created by the same behaviour as the red hypothesis. 
    We can observe, that the evidence scores for higher order hypothesis (red) and random (purple) are similar. 
    Hence HypTrails is not able to distinguish the higher order dependencies.
    For a deeper understanding of the plot, we refer to author to~\cite{DBLP:conf/www/SingerHHS15}. 
    \label{fig:ablation-even-biased-hyptrails}
    }
\end{figure}

First, we will show some additional experiments for our approach. 
\Cref{fig:first-even-biased} shows the result of our model trained on \texttt{first even biased}.
Meaning 90\% of the transitions follow \texttt{first even} behavior and the other 10\% the opposite behavior (\texttt{first odd}).
We can see, even with noise that the model is able to grasp the underlying behavior. 
The Hypothesis, which contains the definite, not noisy behavior (purple line, \texttt{first even}) has the overall lowest average rank per position, closely followed by the hypothesis with the same user behavior (brown line, \texttt{first even biased}). 
For most of the other lines, we can see similar behavior as in \Cref{fig:mixed-higher-order}, except the phenomenon, that the first half of the orange line (\texttt{odd} behavior) has a higher average rank than the second half of the blue line (\texttt{even} behavior). 
Intuitively, these lines should be at the same height.

\subsection{DeepMixedTrails: Higher-order observations}
\begin{figure}
    \centering
    \hspace*{\fill}%
    \begin{subfigure}{0.4\textwidth}
        \includegraphics[width=\textwidth]{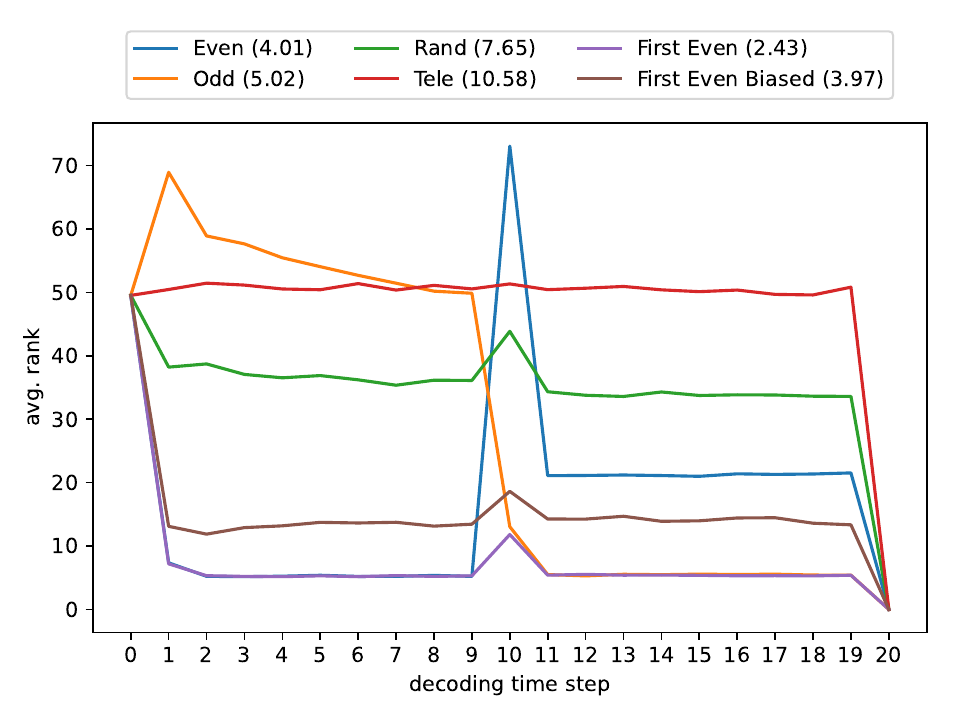}
        \caption{Changing behavior}\label{fig:mixed-higher-order}
    \end{subfigure}\hfill%
    \begin{subfigure}{0.4\textwidth}
        \includegraphics[width=\textwidth]{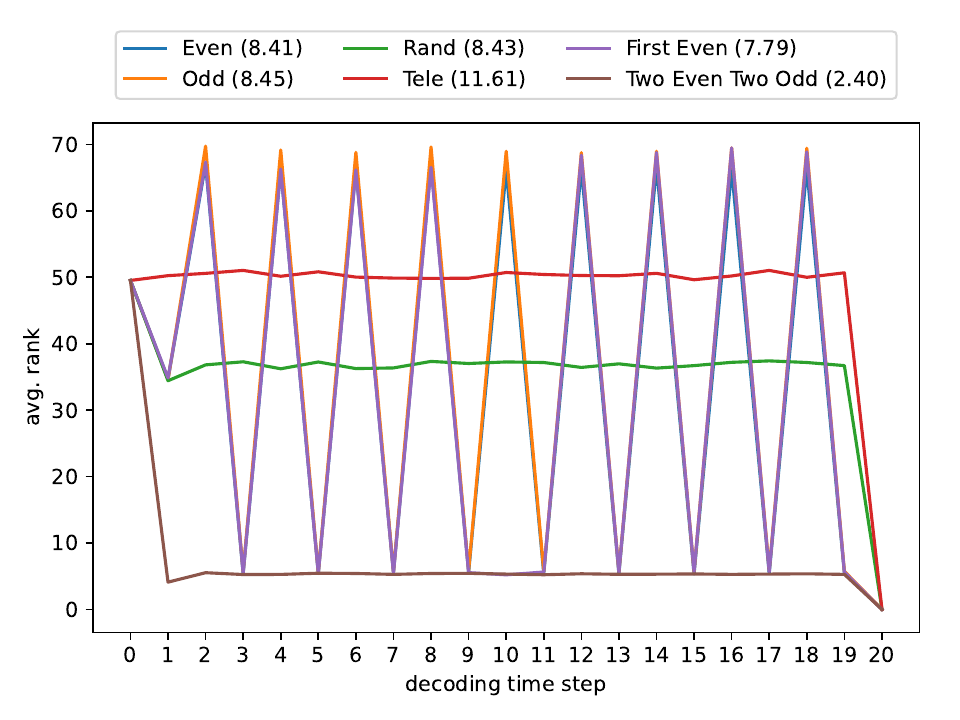}
        \caption{Using higher-order dependencies}\label{fig:mixed-two-two}
    \end{subfigure}%
    \hspace*{\fill}%
    \caption{Visualisation of DeepMixedTrails second setting, where a model is trained on sequences with the same driving force, but containing higher-order dependencies.
    \Cref{fig:mixed-higher-order} shows a model trained on \texttt{first even} observations. 
    \Cref{fig:mixed-two-two} displays the result when training on observations, which transitions twice to even nodes, then two time to an odd node.  
    }
    \label{fig:mixed-appendix}
\end{figure}

The first scenario of DeepMixedTrails is displayed in \Cref{sec:method:mixedtrails} and contains a dataset with different groups of users, each showing different behavior. 
Now, the second scenario assumes that a sequence contains changing transition probabilities, as seen in \Cref{fig:mixed-appendix}.
Here we demonstrate two scenarios. 
\Cref{fig:mixed-higher-order} shows a model trained only on \texttt{first even} observations.
For this, we trained a model on a single group of users with the same higher-order behavior, namely \texttt{first even} bias.
We can see that the respective hypothesis \texttt{first even} has a low rank across all steps.
Hypotheses with first-order dependencies (such as \texttt{even} and \texttt{odd}) have a low rank for half of the sequence, but not in general - as expected.

Furthermore, \Cref{fig:mixed-two-two} shows another higher-order approach, where during training the biased random walker only transitions to even nodes for two steps and then transitions to odd nodes for another two times.
This pattern is then repeated until the end of the sequence.
We can see the expected behavior, since the respective hypothesis has a low rank across all steps (purple line). 
We can observe jumps every two steps for the single modality hypothesis (blue or orange line). 

\begin{figure}
    \centering
    \includegraphics[width=0.6\textwidth]{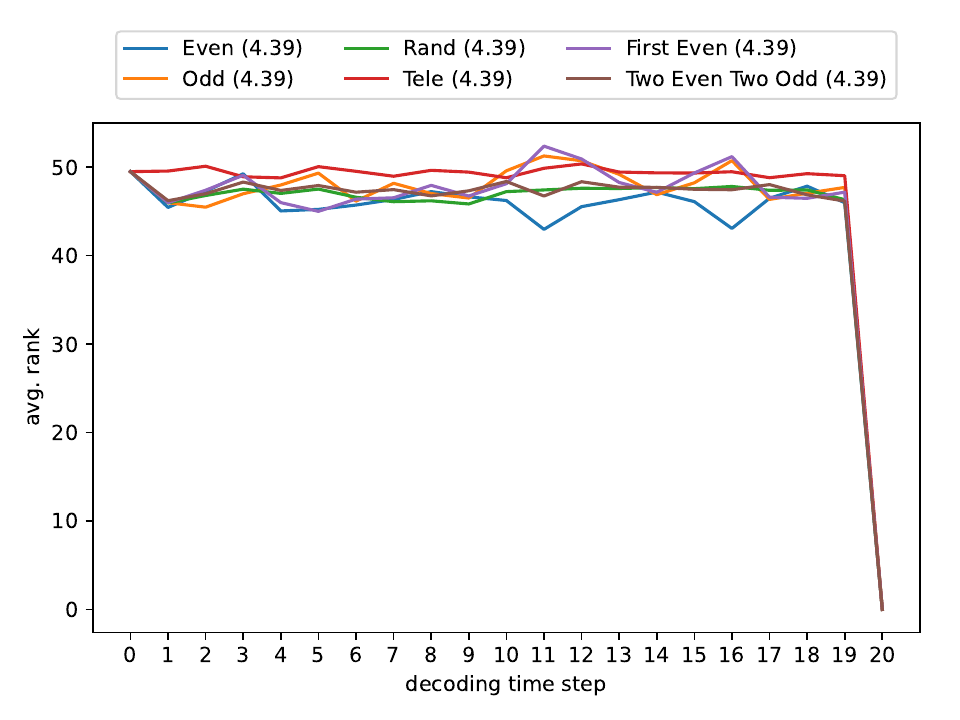}
    \caption{Ablation study: Training a model on teleport behavior. As expected, no hypothesis can explain the underlying data. }
    \label{fig:teleport}
\end{figure}

\subsection{Further Analysis for DeepSubTrails}

\begin{figure}
    \centering
    \hspace*{\fill}%
    \begin{subfigure}{0.49\textwidth}
        \includegraphics[width=\textwidth]{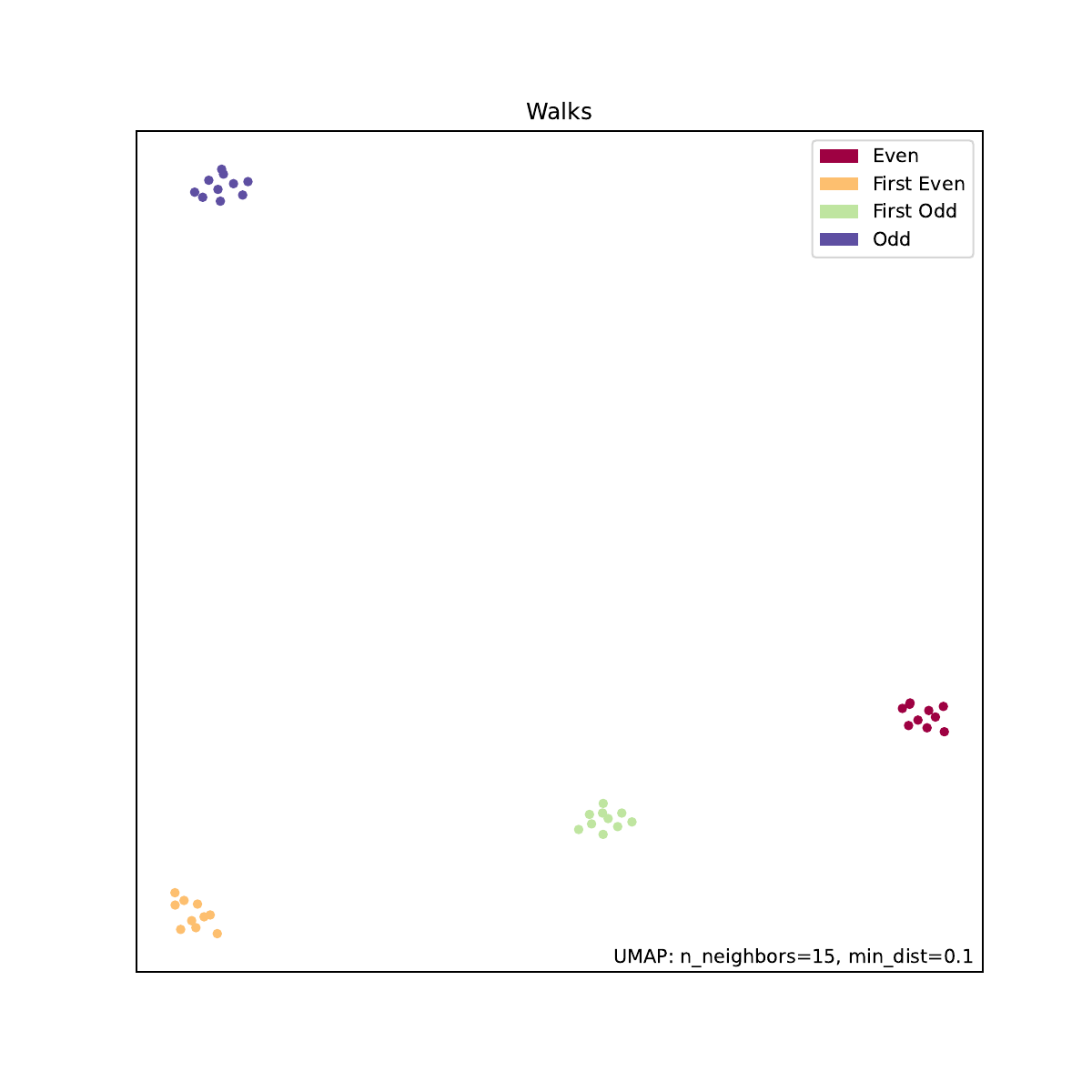}
        \caption{Clustering the walks}
        \label{fig:subtrails-walks}
    \end{subfigure}\hfill%
    \begin{subfigure}{0.49\textwidth}
        \includegraphics[width=\textwidth]{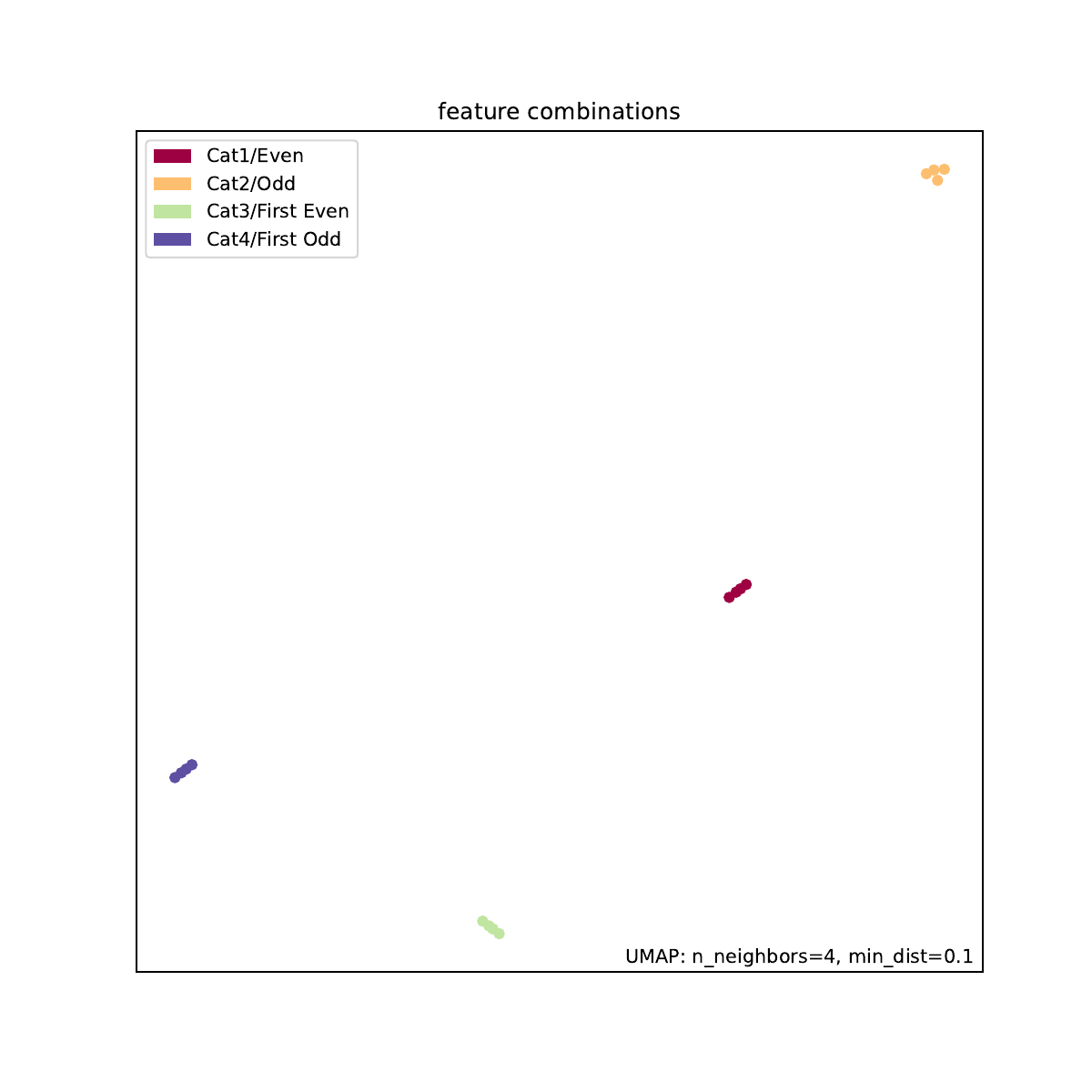}
        \caption{Clustering the features}
        \label{fig:subtrails-features}
    \end{subfigure}%
    \hspace*{\fill}%
    \caption{Analysis of features and walks for our synthetic DeepSubTrails dataset. 
    As in the main manuscript for in \Cref{fig:amzfeatclusters}, we cluster using UMAP and HDBSCAN. 
    We can see that our model is able to distinguish the sequences containing different behavior and the respective cluster neatly.  
    }
    \label{fig:enter-label}
\end{figure}

Furthermore, we show an additional analysis by clustering the loss scores with respect to the features (see \Cref{fig:enter-label}).
The different features, which contain walks with different behaviors, are separable. 
We can observe the same result, when clustering the sequences, as shown in \Cref{fig:subtrails-features}. 

\end{document}